%% file: main.tex
\documentclass[10pt,twocolumn,letterpaper]{article}

\usepackage[pagenumbers]{cvpr} % To force page numbers, e.g. for an arXiv version

\usepackage{caption}
\usepackage{subcaption}
\usepackage{graphicx}
\usepackage{amsmath}
\usepackage{amssymb}
\usepackage{booktabs}
\usepackage{multirow}
\usepackage{bm}
\usepackage{makecell}
\usepackage{marvosym}
\usepackage{shadowtext}
\usepackage{filecontents}
\shadowoffset{3pt}

\input{math_commands.tex}

\usepackage[pagebackref,breaklinks,colorlinks]{hyperref}

\usepackage[capitalize]{cleveref}
\crefname{section}{Sec.}{Secs.}
\Crefname{section}{Section}{Sections}
\Crefname{table}{Table}{Tables}
\crefname{table}{Tab.}{Tabs.}

\def\eg{\emph{e.g.}}
\def\ie{\emph{i.e.}}

\usepackage{pifont}
\usepackage[perpage,symbol*]{footmisc}
\DefineFNsymbols{circled}{{\ding{192}}{\ding{193}}{\ding{194}}
{\ding{195}}{\ding{196}}{\ding{197}}{\ding{198}}{\ding{199}}{\ding{200}}{\ding{201}}}
\setfnsymbol{circled}

\newcommand\myfootnotestyle[1]{\ifcase#1 \or \ding{182}\or \ding{183}\or
\ding{184}\or \ding{185}\or \ding{186}\or \ding{187}%
\or \ding{188}\or \ding{189}\or \ding{190}\or \ding{191}\else *\fi\relax}

\def\cvprPaperID{2527} % *** Enter the CVPR Paper ID here
\def\confName{CVPR}
\def\confYear{2023}

\begin{document}

\title{Universal Backdoor Attacks Detection via Adaptive Adversarial Probe}

\author{
\textbf{Yuhang Wang}$^1$\,,
\textbf{Huafeng Shi}$^1$\,,
\textbf{Rui Min}$^1$\,,
\textbf{Ruijia Wu}$^1$\,,
\textbf{Siyuan Liang}$^2$\,,
\textbf{Yichao Wu}$^1$\,\\
\textbf{Ding Liang}$^1$\,,
\textbf{Aishan Liu}$^3$$^{*}$ \\
$^1$\normalsize SenseTime Research \\
$^2$\normalsize Institute of Information Engineering, Chinese Academy of Sciences \\
$^3$\normalsize NLSDE, Beihang University, Beijing, China \\
}
\maketitle
\begin{abstract}
Extensive evidence has demonstrated that deep neural networks (DNNs) are vulnerable to backdoor attacks, which motivates the development of backdoor attacks detection. Most detection methods are designed to verify whether a model is infected with presumed types of backdoor attacks, yet the adversary is likely to generate diverse backdoor attacks in practice that are unforeseen to defenders, which challenge current detection strategies. In this paper, we focus on this more challenging scenario and propose a universal backdoor attacks detection method named \textit{Adaptive Adversarial Probe} (A2P). Specifically, we posit that the challenge of universal backdoor attacks detection lies in the fact that different backdoor attacks often exhibit diverse characteristics in trigger patterns (i.e., sizes and transparencies). Therefore, our A2P adopts a global-to-local probing framework, which adversarially probes images with adaptive regions/budgets to fit various backdoor triggers of different sizes/transparencies. Regarding the probing region, we propose the attention-guided region generation strategy that generates region proposals with different sizes/locations based on the attention of the target model, since trigger regions often manifest higher model activation. Considering the attack budget, we introduce the box-to-sparsity scheduling that iteratively increases the perturbation budget from box to sparse constraint, so that we could better activate different latent backdoors with different transparencies. Extensive experiments on multiple datasets (CIFAR-10, GTSRB, Tiny-ImageNet) demonstrate that our method outperforms state-of-the-art baselines by large margins (\textbf{\textit{+12\%}}).\footnote{Our codes will be available upon paper publication.}

\end{abstract}

%%%%%%%%% BODY TEXT
\input{src/1-intro}
\input{src/2-relatedwork}
\input{src/3-threatmodel}
\input{src/4-method}

\input{src/5-exp}

\input{src/6-conclu}

%%%%%%%%% REFERENCES
{\small
\bibliographystyle{ieee_fullname}
\bibliography{egbib}
}
\clearpage
\end{document}

%% file: math_commands.tex
\usepackage{amsmath,amsfonts,bm}

\def\eqref#1{equation~\ref{#1}}
\def\1{\bm{1}}

\DeclareMathAlphabet{\mathsfit}{\encodingdefault}{\sfdefault}{m}{sl}
\SetMathAlphabet{\mathsfit}{bold}{\encodingdefault}{\sfdefault}{bx}{n}

%% file: src/1-intro.tex
\section{Introduction}
\label{sec:intro}

\input{resources/fig_1_intro}

DNNs have shown strong potential in various areas including computer vision, natural language processing, and acoustics \cite{krizhevsky2012imagenet, he2016deep,devlin2018bert}. Currently, machine Learning as a Service (MLaaS) platforms have emerged to outsource well-trained deep learning models for developers since it often requires high computational resources for training high-quality DNNs. However, severe security issues exist when using online platforms due to the black-box training process. For example, adversaries could manipulate model behaviors with specific trigger patterns when inference by embedding backdoors \cite{gu2017badnets} into models during training.

To mitigate the threats brought by backdoor attacks, a long line of backdoor attacks detection methods has been proposed \cite{wang2019neural,guo2020towards,wang2020practical,tran2018spectral,gao2019strip,chou2020sentinet}. Generally, the mainstream backdoor detection could be roughly divided into \emph{pre-training} (\ie, whether the training example is poisoned) and \emph{post-training} (\ie, whether the model is infected) detection. Since the training dataset is often hard to access for defenders, this paper primarily focuses on a more practical scenario of post-training detection, \ie, detecting whether a model is injected with the backdoor. Based on the unavailability of poisoned training data, current detection methods often presume prior knowledge of backdoor triggers and focus on detecting specific types of backdoor attacks. For example, \cite{guo2021aeva, gu2017badnets} could effectively detect backdoor attacks with small trigger patterns while failing in large trigger patterns. However, in practice, the adversary is likely to embed diverse backdoor attacks containing different trigger patterns that are unforeseen to defenders (\eg, invisible \cite{nguyen2021wanet} and sample-specific \cite{nguyen2020input}), which highly challenge the generalization of existing backdoor detection methods. 

% on this more challenging scenario
In this paper, we focus on universal post-training backdoor detection against diverse unforeseen backdoor attacks. Specifically, we posit that universal backdoor attacks detection should overcome the challenge of the diverse characteristics of trigger patterns (\ie, sizes and transparencies). Therefore, we propose \emph{Adaptive Adversarial Probe} (A2P) approach, which utilizes adversarial perturbation as a probe to activate model shortcut for backdoor identification. In particular, our approach works in a global-to-local manner, where we adaptively adjust our adversarial probes in the probing regions and budgets to fit the diverse trigger sizes and transparencies brought by different types of unforeseen backdoor attacks. Regarding the probing region, we propose the attention-guided region generation strategy that generates region proposals with different sizes/locations based on the attention of the target model, since trigger regions often manifest higher model activation. Considering the attack budget, we introduce the box-to-sparsity scheduling that iteratively increases the perturbation budget from box to sparse constraint, so that we could better activate different latent backdoors with different trigger transparencies. Extensive experiments on CIFAR-10, GTSRB, and Tiny-ImageNet demonstrate that our A2P achieves promising performance in detecting diverse unforeseen backdoor attacks and outperforms existing baselines by large margins (\textbf{\textit{+12\%}}). Our \textbf{contributions} are:
\begin{itemize}
    \item We propose A2P framework that works in a global-to-local probing manner to detect infected models that may be embedded with diverse unforeseen backdoors.
    \item For the region, we propose the attention-guided region generation for generating different attacking region proposals; for the budget, we introduce the box-to-sparsity scheduling that iteratively increases the budgets from box to sparse constraint.
    \item Extensive experiments demonstrate that our A2P could achieve promising performance on diverse unforeseen backdoor attacks, and outperform others largely.
\end{itemize}

%% file: resources/fig_1_intro.tex
\begin{figure}%[!htb]
\centering
\vspace{-0.1in}
\includegraphics[width=0.9\linewidth]{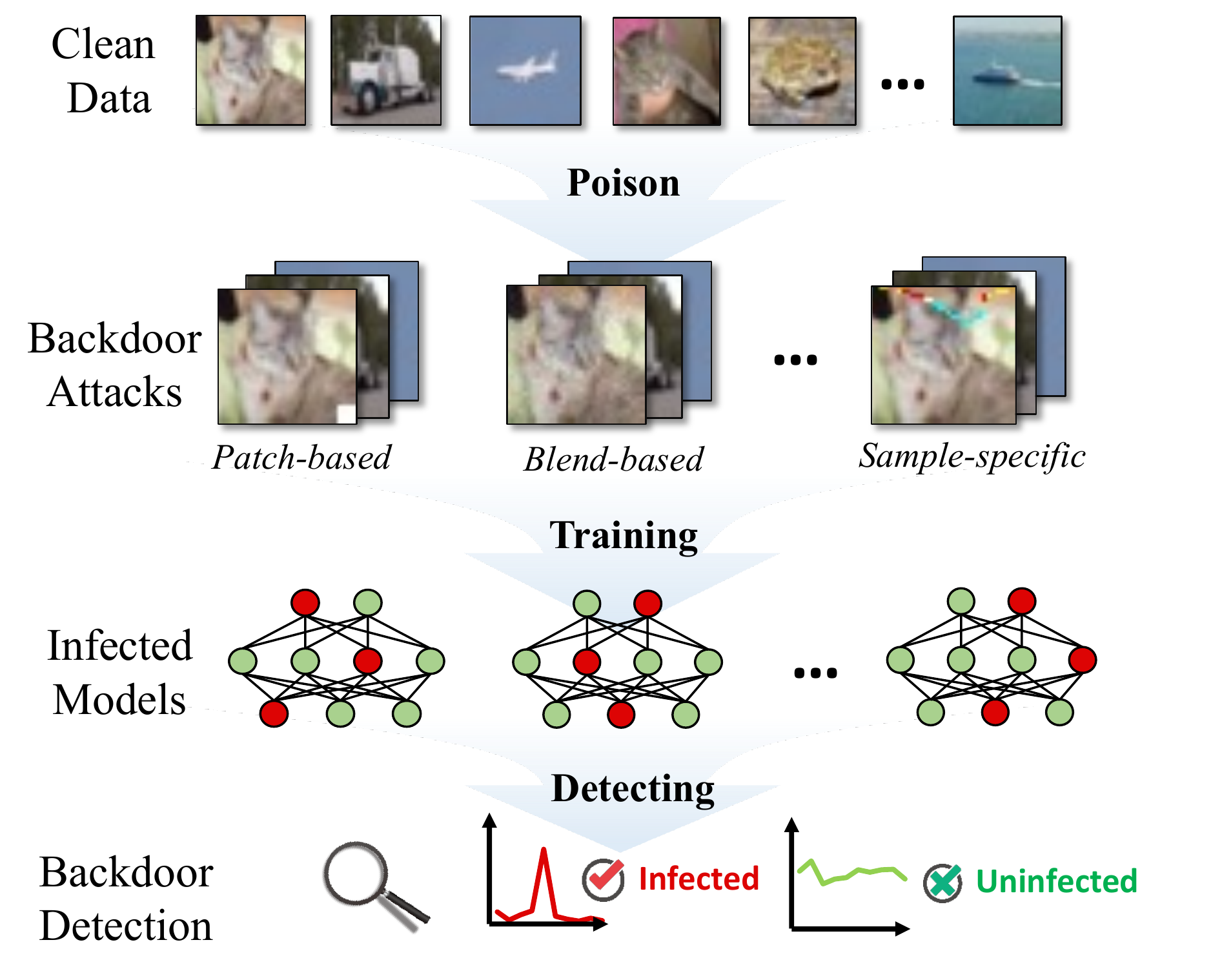}
\caption{Previous backdoor detection methods focus on detecting whether a model is infected with a presumed type of backdoor attack. In this paper, we focus on the more challenging scenario, where defenders aim to identify infected models that might be embedded with diverse types of unforeseen backdoor attacks.}
\vspace{-0.1in}
\label{fig:picture001}
\end{figure}

%% file: src/2-relatedwork.tex
\section{Related Work}
\label{related}
\subsection{Backdoor Attack} 
Backdoor attack mainly affects the training process and forces a mapping between the trigger pattern and the target label. The models embedded with backdoors show malicious behavior when the input image is tampered with a trigger, otherwise, behave normally. \cite{gu2017badnets} first proposed BadNets by sticking a patch-based trigger on the training data and changing their labels to a specific target class (dirty-label attack). Meanwhile, \cite{liu2017trojaning} optimized the trigger pattern and implemented the backdoor attack using transfer learning. However, the patch-based trigger could be detected by humans easily which motivates the researches on designing more stealthy backdoor attacks. \cite{chen2017targeted} poisoned the training dataset with a global pattern and increased trigger transparency to evade human inspection. \cite{nguyen2020input} designed both a mask and trigger generator to generate sample-specific triggers. \cite{nguyen2021wanet} utilized image wrapping to make the poisoned image natural-looking. Besides these dirty-label attacks, other attacks \cite{turner2019label, barni2019new, saha2020hidden, zhao2020clean} considered poisoning the data in the target class without changing the original label (clean-label attack), which further increase the stealthiness of attacks.

% grammarly modified
\subsection{Backdoor Detection} To mitigate backdoor attacks, a long line of detection methods has been proposed. Typically, current backdoor detection could be divided into poisoned dataset detection (\emph{pre-training}) \cite{tran2018spectral, gao2019strip} and backdoor model detection (\emph{post-training}) \cite{wang2019neural}. Since the poisoned dataset is often hard to access, this paper considers a more practical scenario to detect whether a model is embedded with backdoor attacks in the post-training stage. Neural Cleanse \cite{wang2019neural} first identified the shortcut in the infected models and detected the backdoor based on trigger reconstruction. The following work \cite{guo2020towards, wang2020practical,zhu2020gangsweep, qiao2019defending} tried to improve the detection accuracy based on the similar trigger reconstruction framework. Some work even focused on black-box setting \cite{guo2021aeva} with only hard output labels and distinguished backdoor models using peak values in adversarial maps. However, these methods are less effective in detecting large triggers. Recently, studies \cite{kolouri2020universal, xu2021detecting} also used extra classifiers to detect models with more types of backdoor attacks, but they still failed to detect unforeseen backdoor attacks. A concurrent study \cite{wang2022universal} implemented a universal backdoor detection method via MM statistics. However, they only focus on patch-based backdoors and utilize targeted attack to optimize perturbations that inefficiently considers each class as the the target.

In contrast to previous studies that primarily focus on presumed backdoor attacks with prior knowledge, we focus on a more practical scenario, where defenders have no prior presumptions and would face diverse unforeseen backdoor attacks with various trigger sizes and transparencies.

\subsection{Adversarial Attack} 
Adversarial attacks are inputs intentionally designed to mislead deep learning models during inference but are imperceptible to human visions \cite{szegedy2013intriguing,goodfellow2014explaining}. Specifically, the adversarial perturbation $\bm{\delta}_{i}$ for each image $\bm{x}_{i}$ should satisfy
\begin{equation}
f_{\theta}(\bm{x}_{i}+\bm{\delta}_{i}) \neq \bm{y}_{i}, \  \operatorname{ s.t. }\ \|\bm{\delta}_{i}\| \leq \bm{\epsilon} ,
\end{equation}
where $\| \cdot \|$ represents the distance metric ($\ell_{1}$, $\ell_{2}$, or $\ell_{\infty}$-norm), $\bm{y}_{i}$ denotes the ground-truth label for the image, and $\bm{\epsilon}$ represents the perturbation budget. A long line of work has been proposed to attack deep learning models \cite{goodfellow2014explaining,kurakin2018adversarial,Liu2019Perceptual,Liu2020Spatiotemporal}, which could be roughly divided into white-box and black-box attacks based on the access to the target model. In this paper, we use the adversarial attack as a probe to help diagnose whether the model is embedded with backdoor attacks.

\input{resources/fig_2_framework}

%% file: resources/fig_2_framework.tex
\begin{figure*}
\centering
\vspace{-0.1in}
\includegraphics[width=15cm]{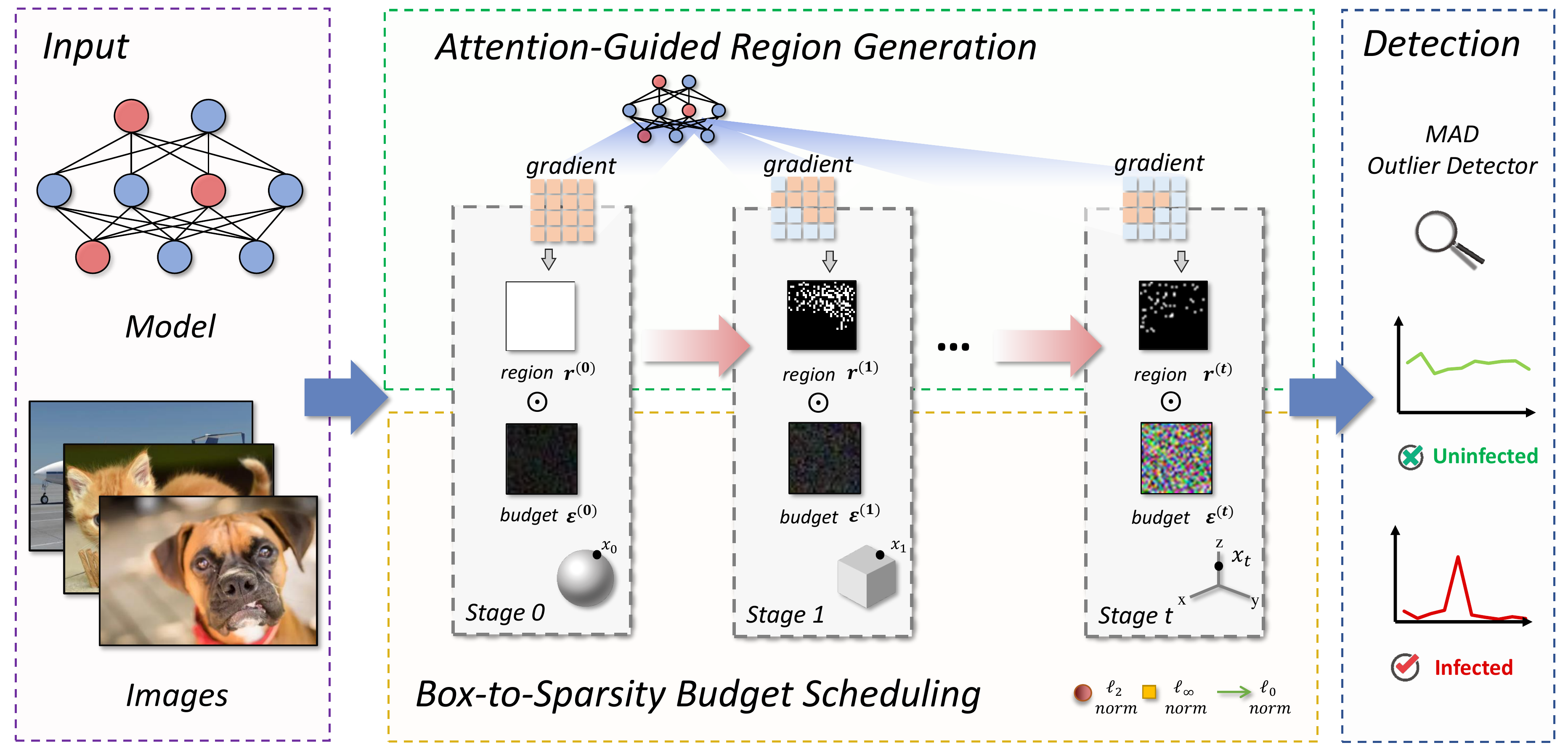}
\caption{Our A2P works in a global-to-local probing manner. In each stage, our attention-guided region generation module first shrinks the probing region based on the gradients of the target model; our box-to-sparsity budget scheduling module then iteratively increases and finds the appropriate probing budget on the attack region; the generated adversarial examples will be finally sent into an outlier detector for subsequent infected model identification.}
\vspace{-0.1in}
\label{fig:picture002}
\end{figure*}

%% file: src/3-threatmodel.tex
\section{Threat Model}
\subsection{Problem Definition}
This paper focuses on image classification task, where a classifier $f_{\theta}$ maps input image $\bm{x}$ $\in$ $\mathbf{X}_{train}$ to label $\bm{y}$ $\in$ $\mathbf{Y}_{train}$. \textbf{{Backdoor attacks}} aim to cheat model $f_{\theta}$ through injecting poisoned data in the \emph{training phase}, so that the infected model would behave maliciously when the inputs are embedded with triggers while behaving normally on clean examples. Specifically, the adversary selects a portion of clean training data $\{\bm x_1,...,\bm x_n\}$ and generates poisoned images $\{\hat{\bm{x}}_{1},...,\hat{\bm{x}}_{n}\}$ for model backdoor training
\begin{equation}
    \hat{\bm{x}}_{i} = \phi (\bm{x}_{i}, \bm{T}),
\end{equation}
\noindent where function $\phi$ is the predefined backdoor attack that generates poisoned images by adding the trigger $\bm{T}$. The models embedded with backdoors would give target label predictions $f_{\theta}(\hat{\bm{x}}_{i}) = \bm{y}_t$ on test images $\hat{\bm{x}}_{i}$ with triggers.

In practice, adversaries are likely to inject \textbf{{different types of unforeseen backdoor attacks}} to escape the backdoor detection. Thus, the trigger pattern should be formalized as $\mathbf{T}=(\bm{\mu},\bm{\sigma})$, where $\bm{\mu}$ is the trigger pattern and $\bm{\sigma}$ is the pattern embedding strategy. Based on that, backdoor trigger injection can be generalized as

\begin{equation}
    \hat{\bm{x}}_{i} =\phi (\bm{x}_{i}, \bm{T})= \phi (\bm{x}_{i},(\bm{\mu},\bm{\sigma})).
\end{equation}

In particular, for the patch-based attack, $\bm{\mu}$ represents the patch trigger, and $\bm{\sigma}$ denotes the binary mask ensuring the pattern's location; for the blend-based attack, $\bm{\mu}$ is the predefined image (\eg, hello kitty and Gaussian noise) and $\bm{\sigma}$ indicates trigger transparency; for the sample-specific attack, $\bm{\sigma}$ denotes the parameters of the trigger generation network $g$, and $\bm{\mu}=g_{\bm{\sigma}}(\bm{x})$ is the trigger pattern for each image.

\subsection{Goals and Challenges}
In this paper, we focus on the \textbf{post-training backdoor detection}, \ie, whether a model is infected by backdoor attacks. In contrast to previous backdoor detection that assumes the model is embedded with a specific type of backdoor attack, we focus on \textbf{a more complex and practical scenario, where defenders have no prior presumptions and would face diverse types of backdoor attacks.} 

We first revisit the classic backdoor detection framework that utilizes reverse engineering to generate the simulated trigger $\tilde{T} = (\tilde{\bm{\mu}},\tilde{\bm{\sigma}})$ for each target label $\bm{y}_{t}$ without accessing the training data. Specifically, the optimization objective using $N$ test images could be formulated as
\begin{equation}
\arg\min_{\tilde{\bm{\mu}},\tilde{\bm{\sigma}}} \sum_{i=1}^{N}\mathcal{L} (f_{\theta} (\phi(\mathbf{x}_{i},(\tilde{\bm{\mu}},\tilde{\bm{\sigma}})) ,\bm{y}_{t}) + \beta \left \| \tilde{\bm{\sigma}} \right \|_{1},
\end{equation}
where $\mathcal{L}(\cdot)$ is the cross-entropy loss and $\tilde{\bm{\sigma}}$ is the mask for reversed trigger $\tilde{\bm{\mu}}$. Such detection framework relies on indispensable assumptions on the backdoor attack type as (1) the reversed trigger to the target class should be small in size, and (2) all images share the same reversed trigger. However, a more practical scenario containing diverse unforeseen backdoor attacks is challenging for defenders due to: \textbf{{Challenge \ding{182}}}: Different backdoor attacks vary in trigger pattern sizes and are often placed in different locations. \textbf{{Challenge \ding{183}}}: Different backdoor attacks tend to manifest different trigger pattern transparencies visually.

Since there exists no prior knowledge of the characteristics of input backdoor attacks, it is highly non-trivial to directly apply existing methods in this scenario, which would degrade their performance and even fail on unforeseen backdoor attacks. 

%% file: src/4-method.tex
\section{Adaptive Adversarial Probe Approach}
\label{method}
%The A2P framework is introduced in this section. We first define the backdoor attack and backdoor detection. Then, we describe the overview of our framework, followed by a detailed introduction and the overall detecting process.

%\subsection{Framework Overview}

% Generally, we aim to implement a universal backdoor detection framework without any prior assumption on backdoor triggers which is applicable for a more practical scenario.
\subsection{Global-to-Local Probing Framework}
Generally, we aim to implement a universal backdoor detection framework without any prior assumption on backdoor triggers which is applicable for a more practical scenario. Since previous work \cite{mu2022adversarial} has revealed the close connection between adversarial perturbations and trigger patterns, it is feasible to utilize adversarial perturbations as a probe to detect latent backdoors. However, directly injecting adversarial perturbations on the whole image may not suffice to detect multiple types of backdoor attacks, since such correlation is highly affected by the sizes and transparencies of trigger patterns. Therefore, we propose the A2P framework by adaptively adjusting the attack region $\bm{r}$ (probe location) and attack budget $\bm{\epsilon}$ (probe strength) in a multi-stage manner to fit various backdoor trigger patterns. In each stage $t$, we adversarially probe the model by $\bm{p}_{i}^{(t)}$ as
\begin{equation}
    \bm{p}_{i}^{(t)} =  \arg\max\limits_{
    \| \bm{r}_i^{(t)} \odot \bm{\delta}_i^{(t)}\|_{\infty} \le \bm{\epsilon}^{(t)}
    } {\mathcal{L}(f_{\theta}(\bm{x}_{i}+\bm{r}_i^{(t)}\odot \bm{\delta}_i^{(t)}), \bm{y}_{i})},
\end{equation}

% \psi(\bm{x}_{i})=

\noindent where $\bm{\delta}_i^{(t)}$ is the adversarial perturbation controlled by budget $\bm{\epsilon}^{(t)}$, and $\bm{r}_i^{(t)} \in \{0,1\}^{W\times H}$ is the region mask. %Notably, the attack budget is fixed within a single stage, while the attack region is sample-specific. 

To link individual stages together, we design a global-to-local probe search strategy that starts by injecting adversarial perturbation on the global image region and gradually shrinks the attack region. At each stage, we generate an individual mask for each image based on our attention-guided region generation strategy and constrain probe locations within the masked area; we utilize the proposed box-to-sparsity budget scheduling strategy to iteratively find the proper probing strength on the attack region; the generated adversarial examples will be sent into an outlier detector for subsequent infected model identification. Our framework is illustrated in Figure \ref{fig:picture002}.
% \mr{to sample-specific?} We further find that our method is effective in detecting sample-specific triggers. This is reasonable since our method inject specific adversarial perturbation to each image which is obviously sample-specific. Therefore we violate the assumption that backdoor triggers are sample-agnostic proposed by many previous detection methods. More experimental details would be discussed in the Section \ref{sec:a & d}.

\input{resources/fig_3_grad-map}

% grammarly modified
\subsection{Attention-Guided Region Generation} 

Rethinking our probe-based detection framework, our objective is to automatically search an optimal region for adversarial probing, which could deviate the clean inputs away from their ground-truth labels to activate the latent backdoor. As challenge \ding{182} stated, different backdoor attacks tend to have different trigger pattern sizes with different locations (\eg, patch-based attacks have patch triggers while blend-based attacks show global semi-transparent triggers). Therefore, the attack region is critical to ensure detection performance and efficiency. 

To address the above challenge, our probing strategy should adaptively adjust the attacking region to better fit the backdoor triggers of different sizes/locations, so that we could activate the latent backdoor. For example, patch-based attacks are sensitive to a local perturbation, while blend-based attacks could be activated by a global range of perturbations (more details could be found in Section \ref{sec:a & d}).

However, directly applying random region search or generation on the whole image is computationally insufficient. We observe that models embedded with backdoors would easily focus on the trigger region due to the internal shortcut, manifested as large model attention near the trigger region (as shown in Figure \ref{fig:grad_guided_proposal}). Therefore, we propose the attention-guided region generation strategy based on the attention (gradients) of the target model to perform region generation. Specifically, given a sample $\bm{x}_{i}$, we generate the corresponding attack region $\bm{r}_{i}^{(t)}$ at each stage $t$ using the attention-guided region generation strategy as
\begin{equation}
   \bm{r}_{i}^{(t)} = {top}_{\lfloor\alpha\times \|\bm{r}^{(t-1)}_{i}\|_{1} \rfloor}(\nabla_{\bm{x}_{i}+\bm{p}_{i}^{(t-1)}} \mathcal{L}(f_{\theta}(\bm{x}_{i}+\bm{p}_{i}^{(t-1)}),\bm{y}_{i})),
\end{equation}
where $\alpha \in \left(0, 1 \right)$ is the scale parameter for region shrinking. ${top}$ generates a binary mask by selecting the region with top $\lfloor \alpha\times \|\bm{r}^{(t-1)}_{i}\|_{1} \rfloor$ gradient values and discarding the rest pixels. Notably, the $\ell_{1}$-norm of $\bm{r}_{i}^{(t)}$ is the same for all samples within the same stage.

In summary, we search for the optimal region of adversarial attack by attention-guided region generation. We first generate global perturbations to simulate blend-based triggers and set $\alpha$ to 0.5; we then shrink the region in half until the region is less than 3\% of the whole image.

\input{resources/fig_4_confusion-matrix}

\subsection{Box-to-Sparsity Budget Scheduling} 

After finding the optimal region, the next step is to inject adversarial probes into the image. However, as challenge \ding{183} indicated, different trigger transparencies also impact the detection performance. Thus, a question emerges: \emph{can we modify the attack budget to an arbitrary value?}

To explore the problem, we conduct adversarial attacks on the whole image region with different attack budgets (as shown in Fig \ref{fig:budget confusion matrix}). On one hand, with a proper attack budget, the output of infected models would skew to the target label, while the clean models show no obvious deviation; on the other hand, if the attack budget is increased to a large value (\eg, 32/255), the model output would collapse due to the excessive attack, which leads to a nearly 100\% Attack Success Rate of adversarial examples (ASR-A). \emph{Therefore, we should find a suitable budget within the specific region}, which satisfies (1) the adversarial probe could successfully activate the latent backdoor and (2) the budget value does not damage model predictions. 

To find a suitable budget that meets the above requirements, we formulate the budget generation process as

\begin{equation}
\begin{aligned}
 \bm{\delta}_{i}^{*(t)} = \arg\max\limits_{\bm{\delta}_i^{(t)}} (\mathcal{L}(f_{\theta}(\bm{x}_{i}+\bm{r}_{i}^{(t)}\odot\bm{\delta}_i^{(t)}),\bm{y}_{i}) + \lambda \| \bm{\delta}_i^{(t)}\|_{\infty}), \\
 \operatorname{ s.t.}\   \left | \frac{1}{N} \sum_{i=1}^{N}\mathbb{I}(f_{\theta}(\bm{x}_{i}+\bm{r}_{i}^{(t)}\odot\bm{\delta}_i^{(t)}) \ne \bm{y}_{i})  - \beta \right| \le \eta ,
\end{aligned}
\end{equation}
where $\mathbb{I}$ is the indicator function, $\beta$ is the attack boundary which ensures a non-excessive attack, $\lambda$ is the balancing parameter, and $\eta$ defines the margin that forces ASR-A to be close to $\beta$.

In order to solve the above optimization problem, we propose the box-to-sparsity budget scheduling strategy. Firstly, at the initial stage $t_{0}$, we set our attacking region $\bm{r}^{(0)}$ as the whole image, while the attacking budget $\bm{\epsilon}^{(0)}$ as 4/255. Obviously, the attack at the initial stage follows the commonly-used setting in the adversarial attack which could be treated as a box-constrained attack. Notably, we regard the ASR-A of the initial stage as the attack boundary $\beta$. As the region shrinks, we incrementally improve the attack budgets to find an optimal value, which satisfy that the ASR-A is close to $\beta$ but not exceed for excessive attacks. %Notably, we dynamically increase the budget bounds according to the margin between current attack and attack boundary which ensures a non-excessive but effective budget.

As the stage continues, we iteratively increase the attack budget while reducing the perturbing region without limitation on the values of the adversarial attacks. From this point of view, our attack area becomes sparser, and our adversarial perturbations are scheduled from the box constraint (\eg, $\ell_2$) to the sparse constraint ($\ell_{0}$). Thus, the budget for stage $t$ should be formulated as
\begin{equation}
    \bm{\epsilon}^{(t)} = \bm{\epsilon}^{(t-1)} + \kappa \times (\beta - \frac{1}{N} \sum_{i=1}^{N}\mathbb{I}(f_{\theta}(\bm{x}_{i}+\bm{p}_{i}^{(t)}) \ne \bm{y}_{i})) ,
\end{equation}
where $\kappa$ controls incremental step size for budget based on the ASR-A at stage $t$. Specifically, we first calculate the ASR-A under $\bm{r}^{(t)}_{i}$ and $\bm{\epsilon}^{(t-1)}$. Then, we increase the budget based on the residuals between the ASR-A and $\beta$ and repeatedly conduct the adversarial attack until finding a budget that leads ASR-A to be close to $\beta$. To be noted, we set the maximum attack times for each stage to 3. 

To sum up, we adjust the attack budget via the box-to-sparsity scheduling which simulates different trigger patterns to ensure the diversity of hidden backdoor activation (see Section \ref{sec:abs} for discussions).

\subsection{Overall Detection Process}

Our overall detection process could be regarded as a global-to-local multi-stage detection pipeline. Starting from the whole region, we generate the global adversarial perturbation with the initial budget. For each stage, our attention-guided region generation strategy first takes model gradients as the criterion to search for the optimal probing region; then, our box-to-sparsity budget scheduling strategy iteratively generates adversarial perturbations and injects into the specific areas; finally, with the generated adversarial examples as inputs, we then calculate the softmax outputs of adversarial examples and utilize MAD to detect the outliers. 

The model is identified as a backdoor model only if the anomaly index is larger than the threshold and the overall detection would stop. Otherwise, the detection would continue and repeat the process until the last stage. The model passing through all the detection stages would be considered as a clean (uninfected) model.

%% file: resources/fig_3_grad-map.tex
\begin{figure}[tb]
\centering
\vspace{-0.1in}
\includegraphics[width=0.9\linewidth]{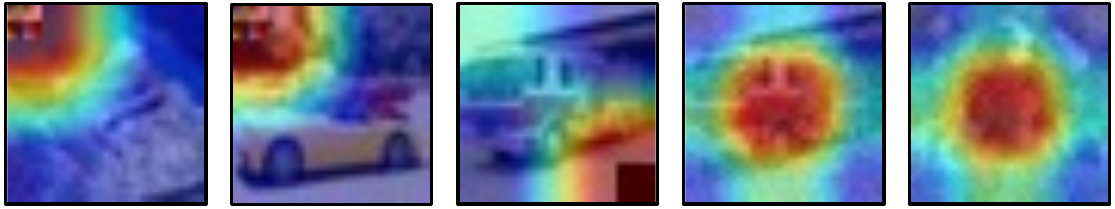}
\caption{Model attention of inputs with triggers using Grad-CAM \cite{selvaraju2017grad} (three images with patch-based triggers and two images with blend-based triggers). The trigger region derives the most attention (gradients) from infected models.}
\vspace{-0.1in}
\label{fig:grad_guided_proposal}
\end{figure}

%% file: resources/fig_4_confusion-matrix.tex
\begin{figure}[tb]
	\centering
	\vspace{-0.1in}
	\begin{subfigure}[]{0.49\linewidth}
        \includegraphics[width=0.95\textwidth, height=0.75\textwidth]{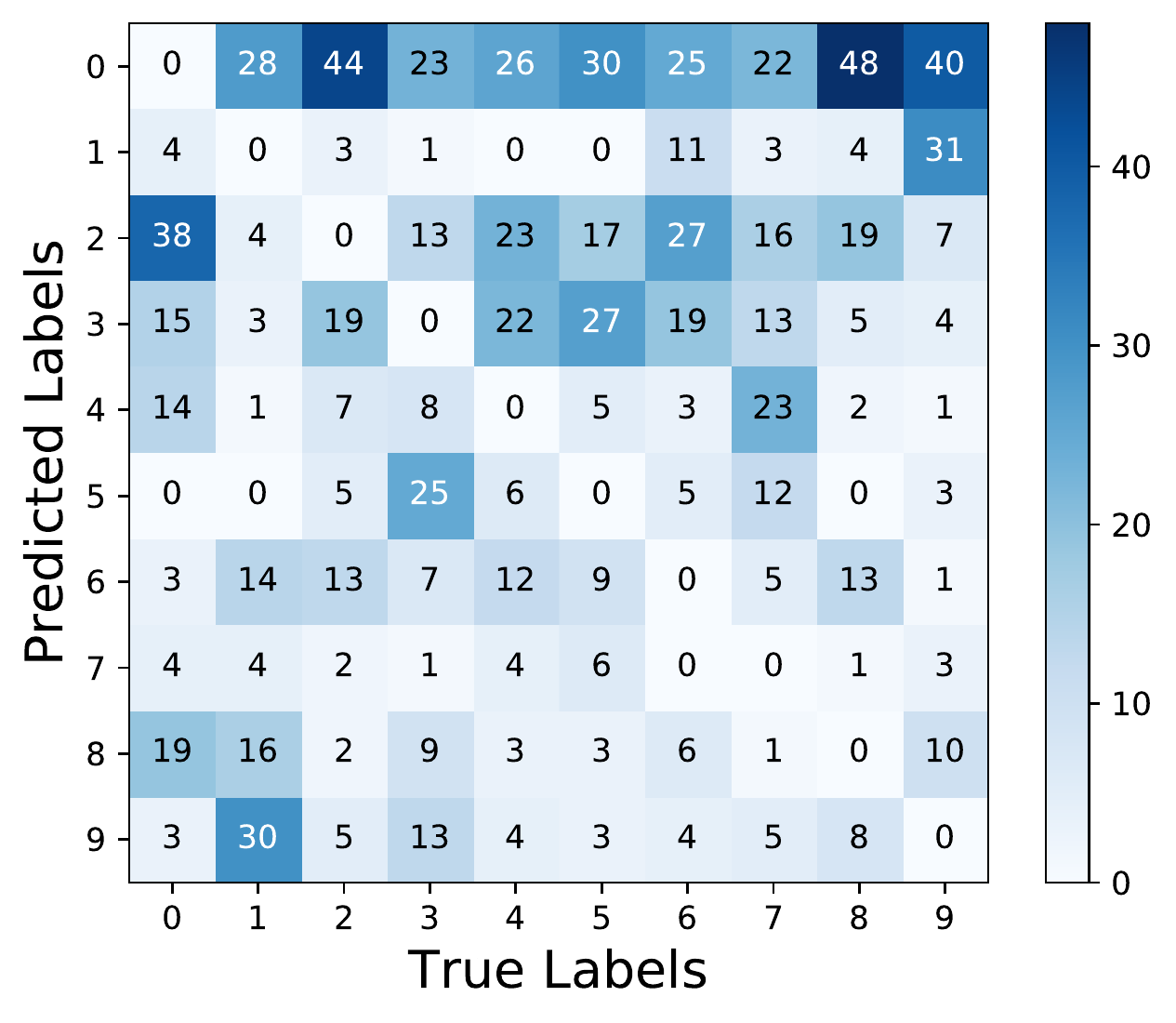}
        \caption{target label = 0}
        \label{fig:backdoor_eps8}
    \end{subfigure}
    \begin{subfigure}[]{0.49\linewidth}
        \includegraphics[width=0.95\textwidth, height=0.75\textwidth]{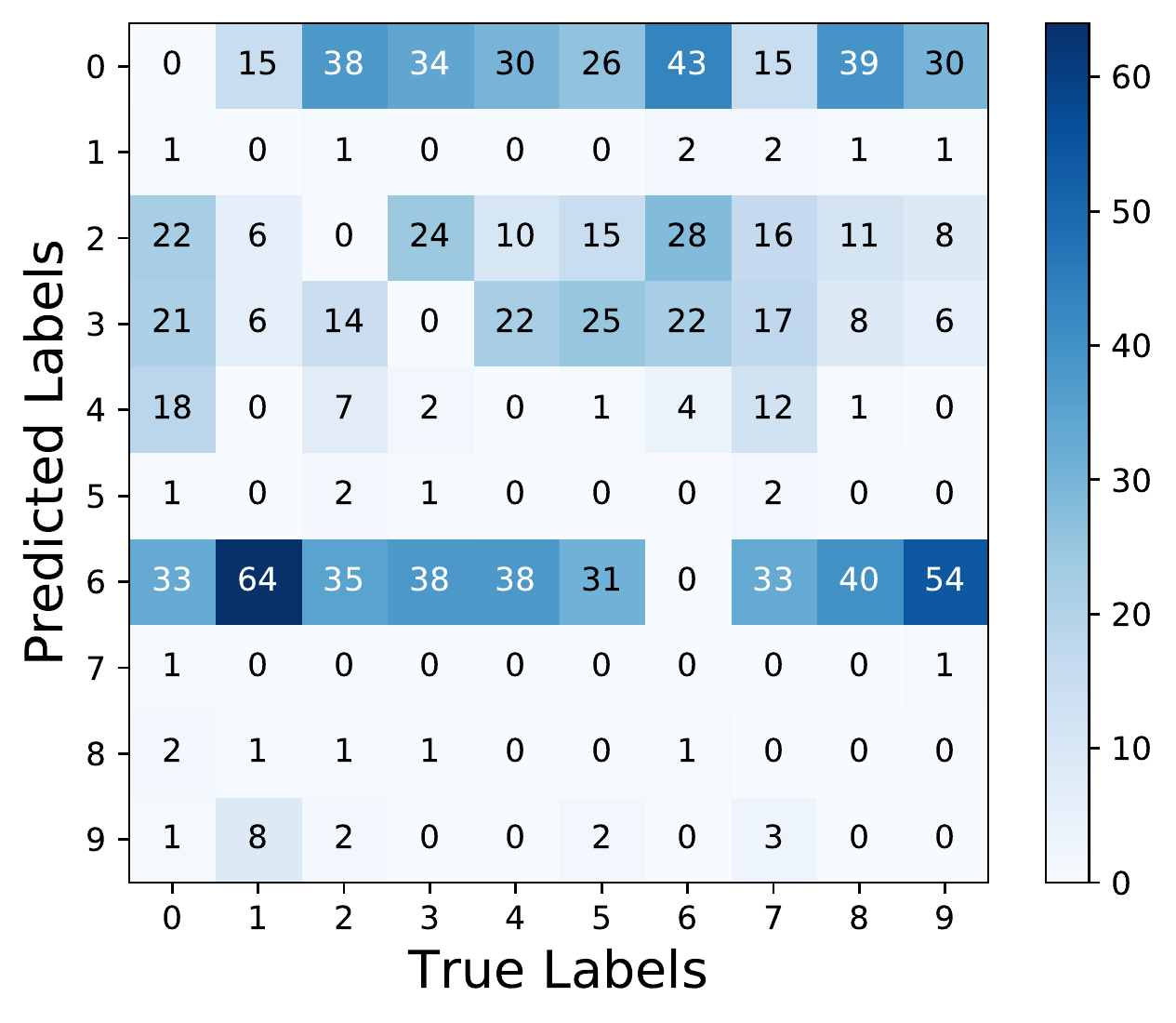}
        \caption{target label = 0}
        \label{fig:backdoor_eps32}
    \end{subfigure}
    
    \begin{subfigure}[]{0.49\linewidth}
        \includegraphics[width=0.95\textwidth, height=0.75\textwidth]{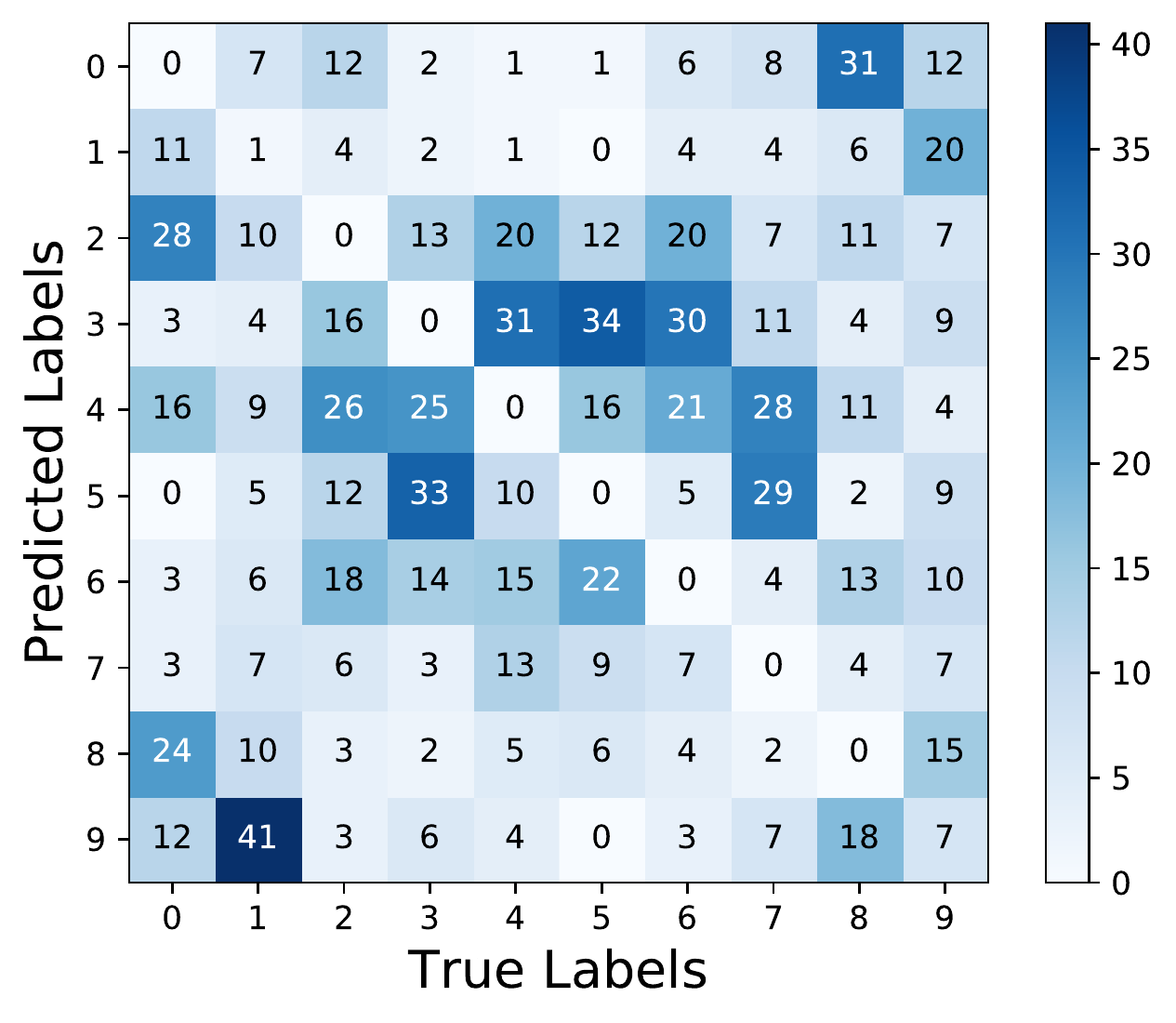}
        \caption{clean}
        \label{fig:clean_eps8}
    \end{subfigure}
    \begin{subfigure}[]{0.49\linewidth}
        \includegraphics[width=0.95\textwidth, height=0.75\textwidth]{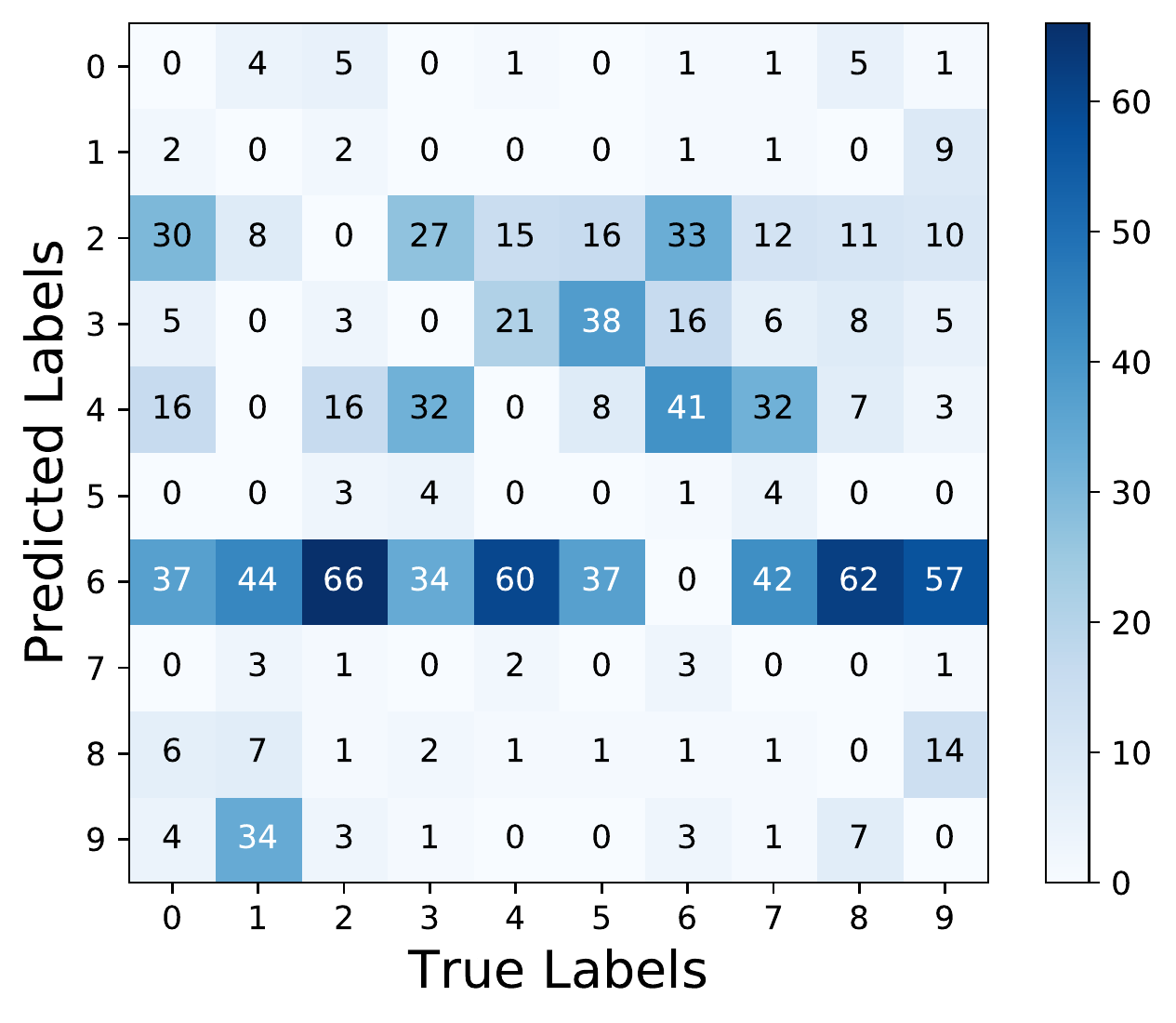}
        \caption{clean}
        \label{fig:clean_eps32}
    \end{subfigure}
	\caption{Confusion matrix of model predictions on global adversarial attacks with different budgets. (a) infected model with small budget (8/255); (b) infected model with large budget (32/255); (c) clean model with small budget; (d) clean model with large budget.}
	\vspace{-0.1in}
	\label{fig:budget confusion matrix}
\end{figure}

%% file: src/5-exp.tex
 \section{Experiments}

\subsection{Experimental Setups}
\label{sec:exp_setting}
% grammarly modified
We first illustrate the experimental setups in this part.

\textbf{Datasets and architectures.} We conduct experiments on image classification tasks using CIFAR-10 \cite{krizhevsky2009learning}, GTSRB \cite{stallkamp2012man}, and Tiny-ImageNet \cite{le2015tiny} datasets. For model architectures, we use ResNet-18 \cite{he2016deep}, VGG19 \cite{simonyan2014very}, DenseNet-161 \cite{huang2017densely}, and MobileNet-V2 \cite{sandler2018mobilenetv2}. 

\input{resources/tab_1_main-exp}

% grammarly modified
\textbf{Backdoor attacks.} We choose 4 commonly-adopted backdoor attacks with different trigger patterns for evaluation including BadNets \cite{gu2017badnets}, Blend \cite{chen2017targeted}, WaNet \cite{nguyen2021wanet}, and Input-aware \cite{nguyen2020input}. For BadNets (patch-based attack), we utilize white square as the backdoor trigger and implement two trigger sizes (\ie, small-scale denoted ``BadNets-s'' and large-scale denoted ``BadNets-l''); for Blend (blend-based attack), we use both Gaussian Noise and Hello Kitty as patterns and implement two trigger transparencies (\ie, low-scale denoted ``Blend-l'' and high-scale denoted ``Blend-h''); for WaNet and Input-aware (sample-specific attack), we use the default settings \cite{nguyen2021wanet,nguyen2020input}. For each attack, we build 60 infected models and 60 benign models evenly distributed over the four architectures on each dataset. Following \cite{guo2021aeva}, we randomly select one target label for each infected model and inject 10\% poisoned samples into training data making the average Attack Success Rate $\ge$ 90\%. %The illustration of poisoned images with triggers are shown in Figure \ref{fig:tot_trigger}.

\textbf{Detection baselines.} We compare our A2P with the state-of-the-art post-training backdoor detection methods Neural Cleanse (NC) \cite{wang2019neural} and DF-TND \cite{wang2020practical}. Specifically, for each dataset, we randomly select 40 test samples evenly from each class to inject adversarial perturbations.

% grammarly modified
\textbf{Implementation details.} For adversarial attacks, we adopt the commonly-used PGD attack \cite{madry2017towards} to perform white-box untargeted attacks. We take 40 steps to optimize adversarial perturbations and set the step size to 0.001. For our MAD detector, we set the threshold as $\tau = 3.5$ for CIFAR-10, $\tau = 6.5$ for GTSRB, and $\tau = 10.0$ for Tiny-ImageNet.

% grammarly modified
\textbf{Evaluation metrics.} Following \cite{kolouri2020universal}, we use \textit{The Area under Receiver Operating Curve} (AUROC) and \textit{Detection Accuracy} (ACC) to evaluate the detection performance on specific types of attacks. We also report \textit{Average Attacks} to calculate the average detection ACC on several types of backdoor attacks. For each metric, higher values mean better performance of backdoor detection.

\subsection{Comparison with Other Baselines}

We first compare A2P with other backdoor detection methods on different attacks. As shown in Table \ref{table:compare}, our A2P framework achieves significantly higher values on \textit{Average Attacks} than others, which demonstrates the overall better performance across different attacks and datasets. We could draw several \textbf{conclusions} below.%In particular, our performance is obviously better than other methods under some specific trigger patterns (\ie, large-size trigger and high-transparency trigger) and is overall better across different attack methods and datasets. 
% \begin{itemize}

(1) For \emph{patch-based attacks (BadNets)}, A2P exhibits stable detection ability against attacks with different trigger sizes, while NC and DF-TND turn out to show weak performance on large triggers (BadNets-l). We attribute this to the particular searching strategy of A2P, which could well fit backdoor triggers with different sizes.

(2) For \emph{blend-based attacks (Blend)}, A2P achieves higher \emph{ACC} and \emph{AUROC} than other baselines across three different datasets. We also notice that A2P is stable against blend-based triggers with different transparencies and achieves an overall detection accuracy of over 99\%. We will further explore the detection stability of A2P with different trigger transparencies in Section \ref{sec: rigorous}.  

(3) For \emph{sample-specific attacks (WaNet and Input-aware)}, we found that our proposed A2P achieves the highest performance in almost all cases across the datasets. However, we should also notice that all methods show comparatively weak detection ability on this type of backdoor attack, which indicates the strong attacking ability of the generated triggers for each specific image.

(4) To better illustrate the general detection performance over different types of backdoor attacks, we also report the \textit{Average Attacks} values, which demonstrate that A2P exhibits significantly better performance across different datasets and outperforms an average of \textbf{\textit{+12\%}} \textit{Average Attacks} compared to baselines.

(5) Apart from \emph{ACC}, we also observe that the overall \emph{AUROC} of A2P is higher than others with an average value of \textbf{0.958}. Such results demonstrate that the high backdoor detection performance (ACC) of A2P does not sacrifice the performance on uninfected models.

To sum up, A2P achieves the best performance compared to existing backdoor detection approaches on detecting diverse unforeseen backdoor attacks across different settings, especially for some invisible attacks (\eg, Blend attack).

\subsection{Detection on More Rigorous Scenarios}
\label{sec: rigorous}
In this section, we further investigate the detection performance of our A2P in more rigorous settings.

\textbf{Different trigger sizes.} We first evaluate our A2P on triggers with different sizes. Specifically, we utilize BadNets (white square as the trigger) with different sizes ranging from 2 $\times$ 2 to 14 $\times$ 14, and we train 24 infected models for each size on CIFAR-10 using ResNet-18. As shown in Figure \ref{fig:acc_abliation}(a), our A2P shows the best performance on triggers with different sizes. However, NC decreases significantly and shows less robustness as the trigger size is larger than 12 $\times$ 12; DF-TND is comparatively stable, yet it still falls behind compared to our A2P. More specifically, our A2P remains effective with ACC $\ge$ 62.5\% even if the trigger size expands to the exaggerated size 14 $\times$ 14, which takes almost 20\% of the whole image. The results demonstrate that A2P is stable to trigger sizes. 

\input{resources/fig_5_trigger-size-and-transparency}

\textbf{Different trigger transparencies.} We then evaluate our A2P on triggers with different transparencies. Specifically, we use Blend attack on CIFAR-10 with trigger transparencies from 0.7 to 0.95. For each trigger transparency, we train 24 infected models using ResNet-18. Figure \ref{fig:acc_abliation}(b) shows that A2P remains effective against all trigger transparencies with ACC $\ge$ 66.67\%, while NC and DF-TND perform worse when trigger transparency is high (0.95) or low (0.7).

\textbf{Multiple triggers within a single image.} We also consider a setting where multiple triggers are simultaneously injected into a single image for training/testing. Specifically, we generate the backdoor trigger by randomly modifying pixels within a $3 \times 3$ area at four corners. Experimental results reveal that the accuracy of A2P would still be stable ($ACC \ge 97.5\%$) as the trigger number increases.

\subsection{Ablation Studies}
\label{sec:abs}

% grammarly modified
\textbf{Attention-guided search.} To evaluate the effectiveness of our attention-guided strategy, we take the random search as a comparison, where we use 40 BadNets models and 40 clean models trained on CIFAR-10 using ResNet-18. Specifically, we shrink the region with the scale parameter set to 0.5 and utilize the box-to-sparsity budget scheduling. As shown in Figure \ref{fig:mask_study}(a), our attention-guided strategy achieves a higher average detection ACC with a large margin compared to the random strategy. Figure \ref{fig:mask_study}(b) shows the results on clean models, where our attention-guided region generation manifests more stability than random search.

\textbf{Box-to-Sparsity budget scheduling.} We then compare our budget scheduling strategy with ``Conservative'' that increases budgets by 2/255 and ``Radical'' that increases budgets exponentially. We use 40 infected models by BadNets and 40 clean models on CIFAR-10 using ResNet-18. As shown in Figure \ref{fig:budget study}, we could observe that our box-to-sparsity budget scheduling achieves the highest average detection ACC compared to other baselines, and also show better performance on clean models.

\textbf{Sample numbers in each class.} Since we utilize the tendency of softmax output on samples in the infected labels for detection, we further study the influence of sample numbers in each class. With the increasing of sample numbers in each class, our A2P shows better detection performance. Importantly, our A2P remains effective with ACC $\ge 90\%$ even if the number of samples in each class reduces to 5.

\input{resources/fig_6_region-study}
\input{resources/fig_7_budget-study}
\input{resources/fig_8_relationship-adv-trigger}

\subsection{Analysis and Discussion}
\label{sec:a & d}
In this section, we provide more studies and analyses to better understand our A2P framework. 

\textbf{Adversarial perturbations v.s. backdoor triggers.} We here study the relationship between adversarial perturbations and different triggers in terms of attack regions/budgets to better understand our framework. Specifically, we select 10 clean models, 10 infected models by BadNets, and 10 infected models by Blend. All models are trained on CIFAR-10 using ResNet-18.

We first study the perturbation region on backdoor detection, where we select the bottom right square corner of images to perform PGD attacks bounded with a fixed budget (8/255). The size changes from $2 \times 2$ to $32 \times 32$. The BadNets triggers are also placed at the bottom right corner. As shown in Figure \ref{fig:region_budget}(a), we observe that (1) for Blend, the detection performance increases as the region size improves and the infected model could be easily detected when the region size is larger than $16 \times 16$; and (2) for BadNets, the anomaly index of infected models remains comparatively low. We conjecture it is due to the incorrect adversarial perturbation budget. We then investigate the perturbation budget on backdoor detection, where we choose 6 budgets ranging from 2/255 to 64/255 in terms of $\ell_{\infty}$-norm with fixed attacking region size ($2\times2$) on the bottom right corner. As shown in Figure \ref{fig:region_budget} (b), we observe that (1) BadNets attack is gradually activated as the budget increases; however (2) the Blend attack is difficult to activate.

Thus, we could draw the conclusion that adversarial perturbations could be simulated as backdoor triggers for backdoor detection only if the region and budget are appropriate.

\input{resources/tab_2_plugins}

\textbf{Backdoor elimination using A2P.} Since our generated probes manifest strong similarities with triggers, we therefore explore whether our probe could be utilized for backdoor elimination via model fine-tuning. Following \cite{wang2019neural}, we select 10\% samples from CIFAR-10 training set and choose 20\% of them to add adversarial probes using A2P. \emph{We then fine-tune infected models for only 1 epoch.} As shown in Table \ref{tab:unlearn}, our A2P could effectively eliminate the latent backdoors by reducing the average attack success rate of backdoor attacks to $<$ 2\% with limited ACC drop ($<$ 2.8\%).

%% file: resources/tab_1_main-exp.tex
\begin{table*}[]\tiny
\vspace{-0.15in}
\caption{Backdoor attacks detection results (\emph{ACC}, \emph{AUROC}) on three datasets. We also report \emph{Average Attacks} that measures the average detection performance across all backdoor attacks. For all metrics, higher values indicate better performance.}
\resizebox{\textwidth}{!}{

\centering
\renewcommand\arraystretch{1}

\begin{tabular}{cc|l|ll|ll|ll}
% \toprule
\hline
\multirow{3}{*}{Attack} & \multirow{3}{*}{} 
& \multicolumn{1}{c|}{\multirow{3}{*}{Method}} & \multicolumn{6}{c}{Detection Results}     \\ 
\cline{4-9} &   
& \multicolumn{1}{c|}{}  
& \multicolumn{2}{c|}{CIFAR-10} 
& \multicolumn{2}{c|}{GTSRB}
& \multicolumn{2}{c}{Tiny-ImageNet}     \\ 
\cline{4-9} &
& \multicolumn{1}{c|}{}
& \multicolumn{1}{c}{ACC(\%)} & \multicolumn{1}{c|}{AUROC} & \multicolumn{1}{c}{ACC(\%)} & \multicolumn{1}{c|}{AUROC} & \multicolumn{1}{c}{ACC(\%)} &\multicolumn{1}{c}{AUROC}  \\ 

\hline
% \midrule

\multirow{6}{*}{BadNets}
& \multirow{3}{*}{-s}   
& NC  & 90.000 & 0.945 & \textbf{100.000} & 0.978 & 63.333 & 0.851  \\
&
& DF-TND & \textbf{98.333} & \textbf{0.999} & \textbf{100.000} & \textbf{1.000} & \textbf{78.333} & \textbf{0.885} \\
& 
& A2P (Ours) & 96.667 & 0.986 & 98.333 & 0.998 & 73.333 & 0.879 \\ 
\cline{2-9}

& \multirow{3}{*}{-l} 
& NC & 73.333 & 0.865 & 91.667 & 0.949 & 18.333 & 0.403 \\
& 
& DF-TND & 91.667 & 0.974 & 90.000 & 0.971 & 41.667 & 0.732 \\
&
& A2P (Ours) & \textbf{93.333} & \textbf{0.976} & \textbf{93.333} & \textbf{0.992} & \textbf{48.333} & \textbf{0.742} \\ 

\hline
% \midrule
\multirow{6}{*}{Blend} 
& \multirow{3}{*}{-h} 
& NC & 58.333 & 0.834 & 93.333 & 0.961 & 80.000 & 0.936 \\
&
& DF-TND & 56.667 & 0.799 & 58.333 & 0.784 & 51.667 & 0.792 \\
&
& A2P (Ours) & \textbf{98.333} & \textbf{0.995} & \textbf{100.000} & \textbf{1.000} & \textbf{98.333} & \textbf{0.978} \\ 
\cline{2-9}

& \multirow{3}{*}{-l}
& NC & 96.667 & 0.981 & 91.667 & 0.958 & 96.667 & 0.986 \\
&
& DF-TND & 58.333 & 0.801 & 61.667 & 0.809 & 50.000 & 0.679 \\
&
& A2P (Ours) & \textbf{98.333} & \textbf{0.990} & \textbf{100.000} & \textbf{1.000} & \textbf{100.000} & \textbf{0.991}  \\

\hline
\multirow{3}{*}{WaNet}
& \multirow{3}{*}{}
& NC & 86.667 & 0.910 & 40.000 & 0.711 & 71.667 & 0.874 \\
&
& DF-TND  & 66.667 & 0.825 & 63.333 & 0.815 & 68.333 & 0.841 \\
&
& A2P (Ours) & \textbf{90.000} & \textbf{0.969} & \textbf{88.333} & \textbf{0.989} & \textbf{78.333} & \textbf{0.929} \\ 

\hline
\multirow{3}{*}{Input-aware}
& \multirow{3}{*}{}
& NC & 20.000 & 0.531  & 66.667  & 0.878  & 6.6667  & 0.630 \\
&
& DF-TND & \textbf{43.333} & 0.681  & 60.000  & 0.843 & \textbf{18.333} & 0.638 \\
&
& A2P (Ours) & 36.667 & \textbf{0.830} & \textbf{74.000} & \textbf{0.968} & 13.333 &  \textbf{0.691} \\

\hline
\multicolumn{1}{l}{\multirow{3}{*}{\textit{Average Attacks}}} 
& \multicolumn{1}{l|}{}
& NC & 70.833 & 0.844 & 80.556 & 0.906 & 56.111 & 0.780 \\
\multicolumn{1}{l}{}
& \multicolumn{1}{l|}{}
& DF-TND & 69.167 & 0.847 & 72.222 & 0.870 & 51.389 & 0.761 \\
\multicolumn{1}{l}{}
& \multicolumn{1}{l|}{} 
& A2P (Ours) & \textbf{85.556} & \textbf{0.958} & \textbf{92.333} & \textbf{0.991} & \textbf{68.611} & \textbf{0.868} \\

\hline
\end{tabular}
}
\label{table:compare}
\end{table*}

%% file: resources/fig_5_trigger-size-and-transparency.tex
\begin{figure}[tb]
\centering
\vspace{-0.1in}
\begin{subfigure}[]{0.49\linewidth}
    \includegraphics[width=\textwidth]{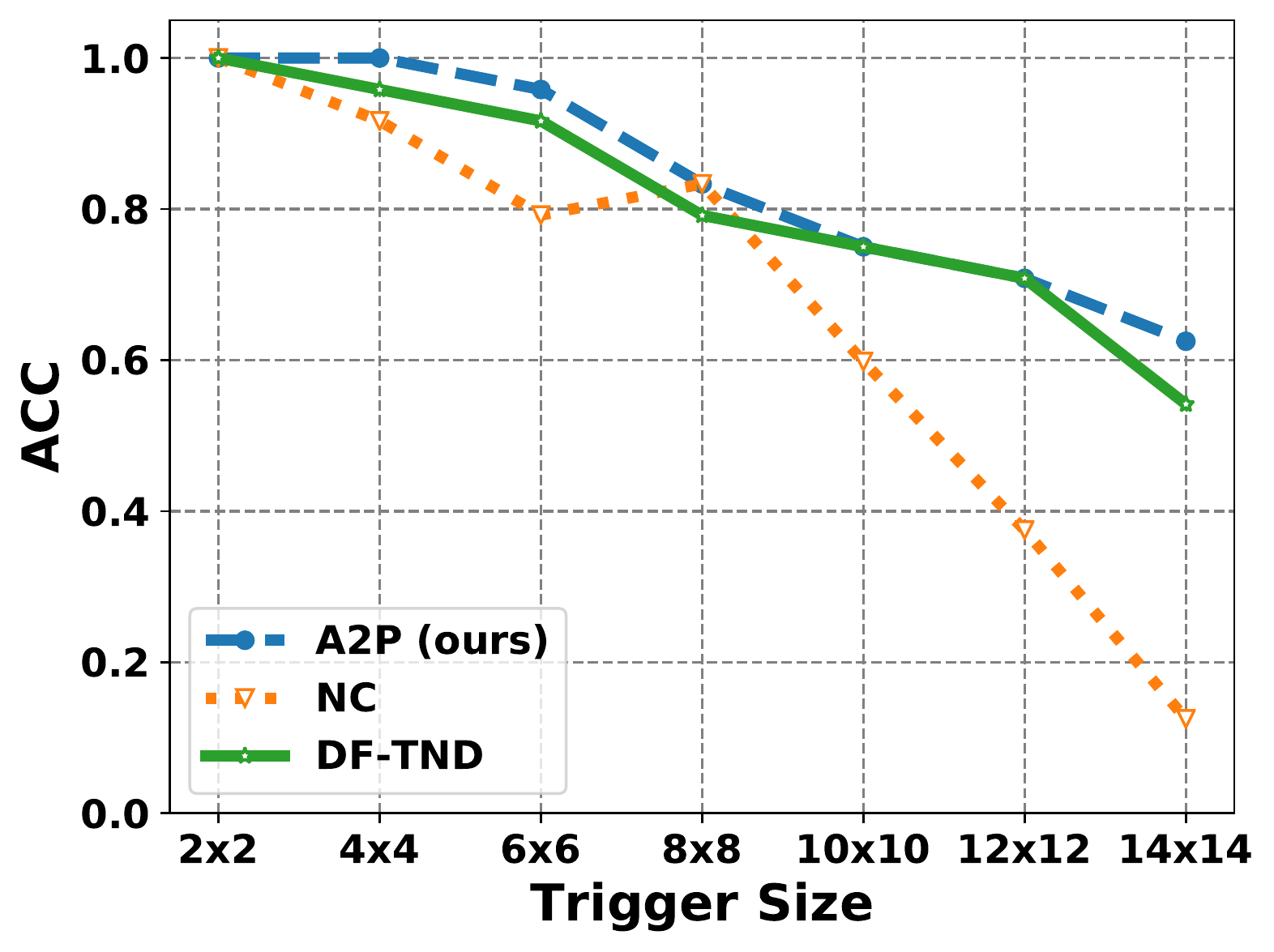}
    \caption{}
    \label{fig:trigger_size}
\end{subfigure}
\begin{subfigure}[]{0.49\linewidth}
    \includegraphics[width=\textwidth]{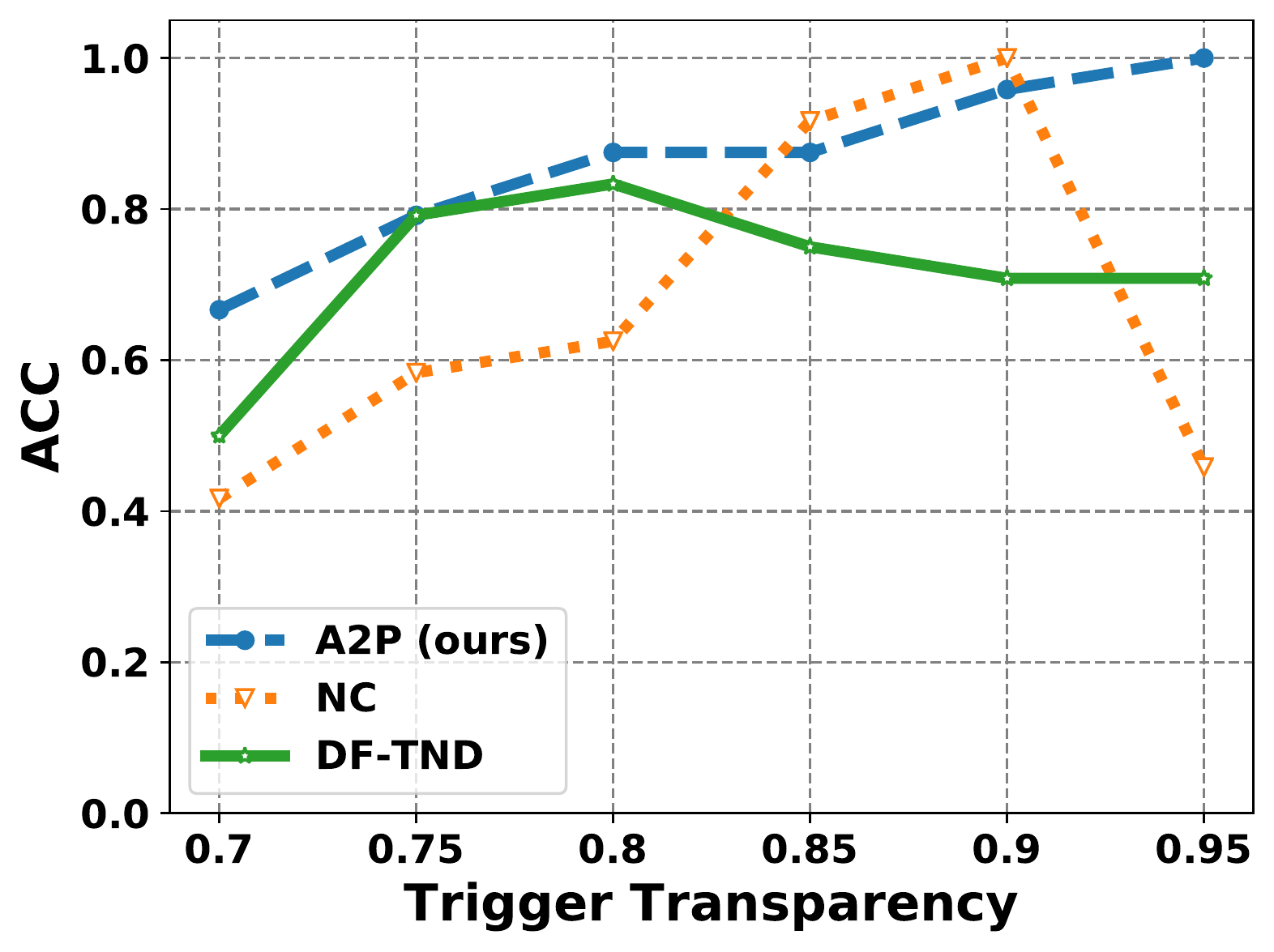}
    \caption{}
    \label{fig:trigger_transparency}
\end{subfigure}
\caption{Detection performance with different trigger patterns on CIFAR-10: (a) trigger sizes and (b) trigger transparencies.}
\vspace{-0.1in}
\label{fig:acc_abliation}
\end{figure}

%% file: resources/fig_6_region-study.tex
\begin{figure}[tb]
\centering
\vspace{-0.1in}
\begin{subfigure}[h]{0.49\linewidth}
    \includegraphics[width=\textwidth]{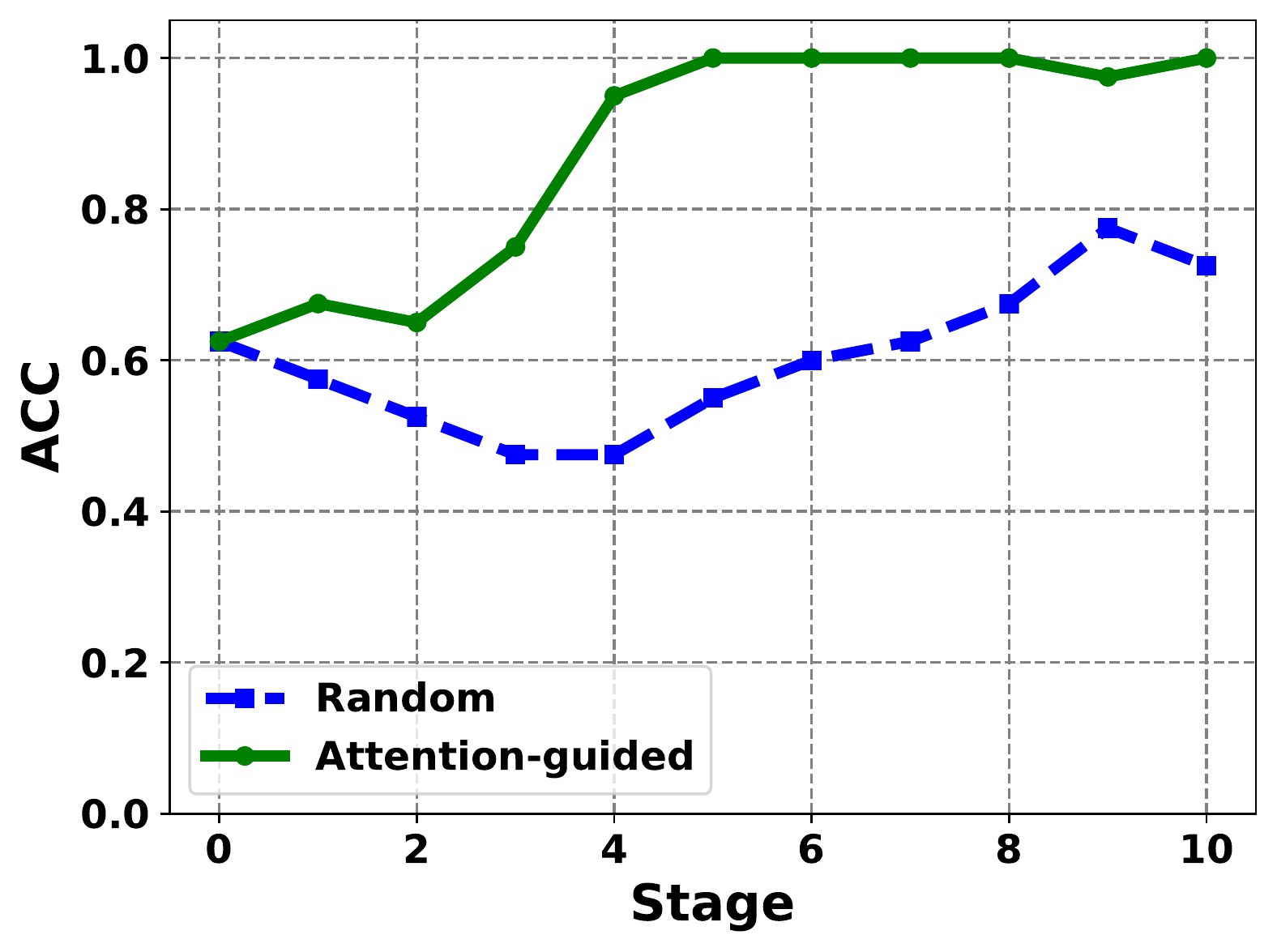}
    \caption{}
    \label{fig:generation_comparison_backdoor}
\end{subfigure}
\begin{subfigure}[h]{0.49\linewidth}
    \includegraphics[width=\textwidth]{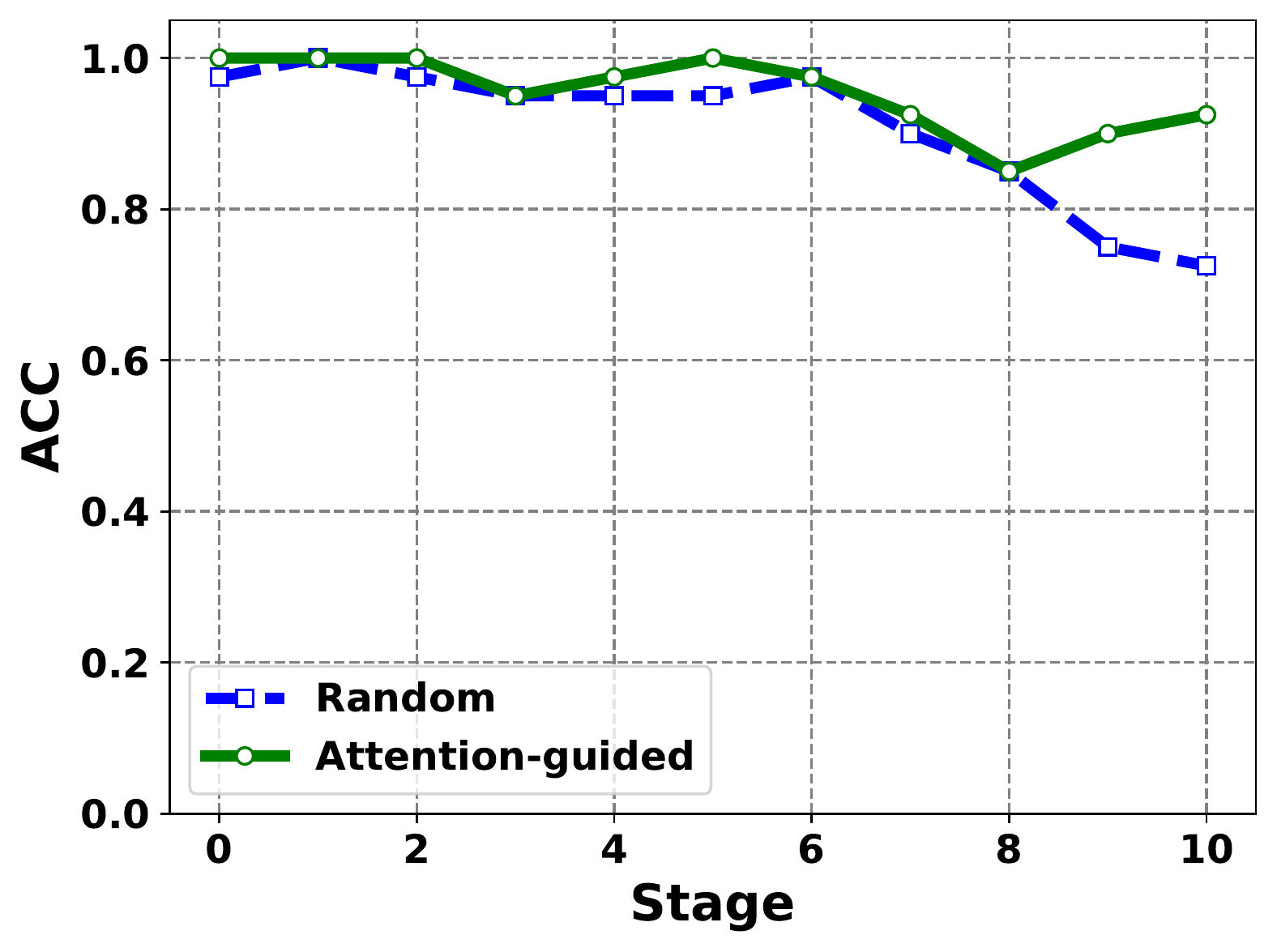}
    \caption{}
    \label{fig:generation_comparison_clean}
\end{subfigure}
\caption{Comparison between different region generation strategies. (a): infected models by BadNets, and (b): clean models.}
\vspace{-0.1in}
\label{fig:mask_study}
\end{figure}

%% file: resources/fig_7_budget-study.tex
\begin{figure}[tb]
    \centering
    % \vspace{-0.1in}
    \begin{subfigure}[Infected models]{0.49\linewidth}
        \includegraphics[width=\textwidth]{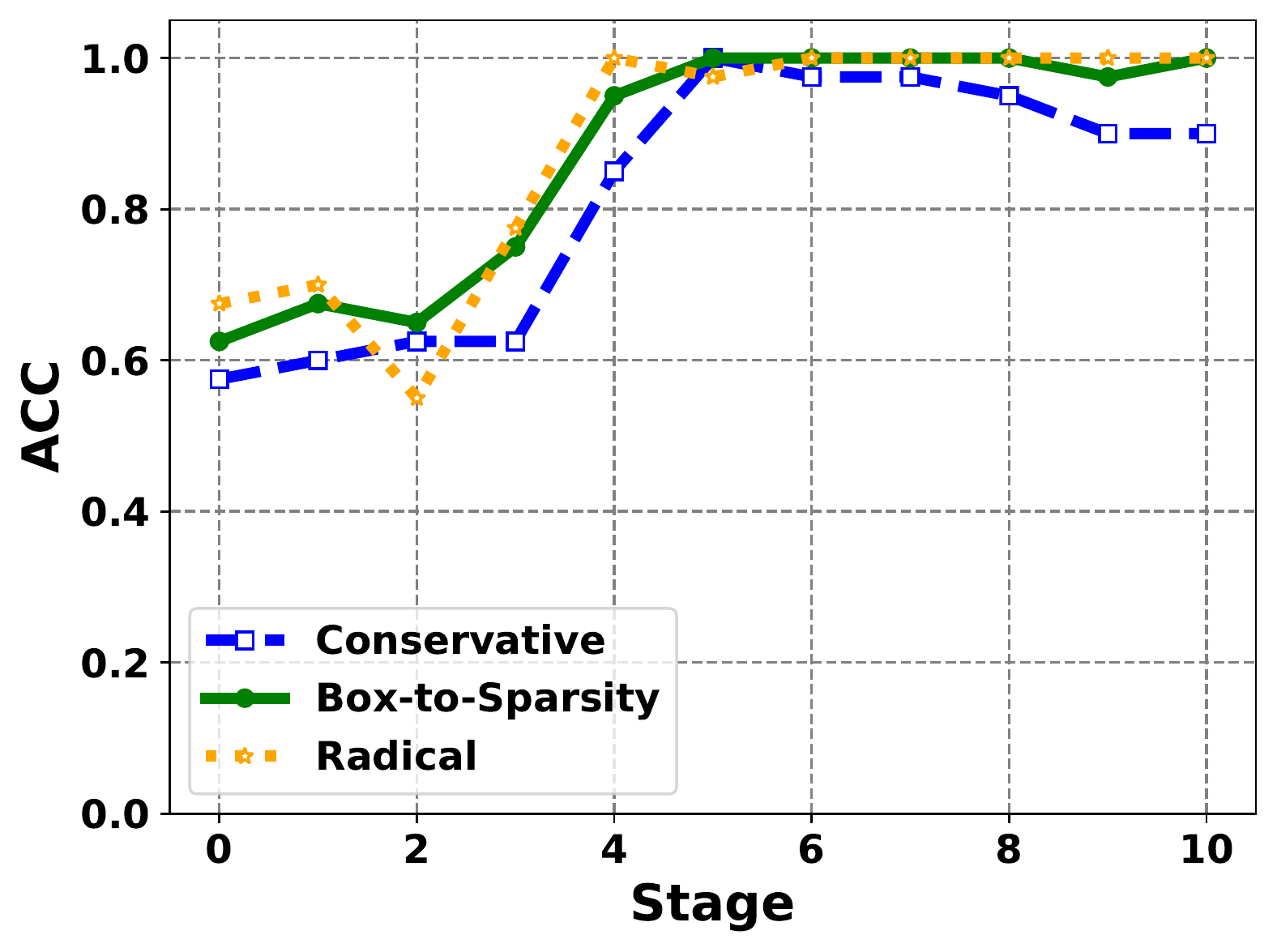}
        \caption{}
        \label{fig:schedule_comparison_backdoor}
    \end{subfigure}
    \begin{subfigure}[Clean models]{0.49\linewidth}
        \includegraphics[width=\textwidth]{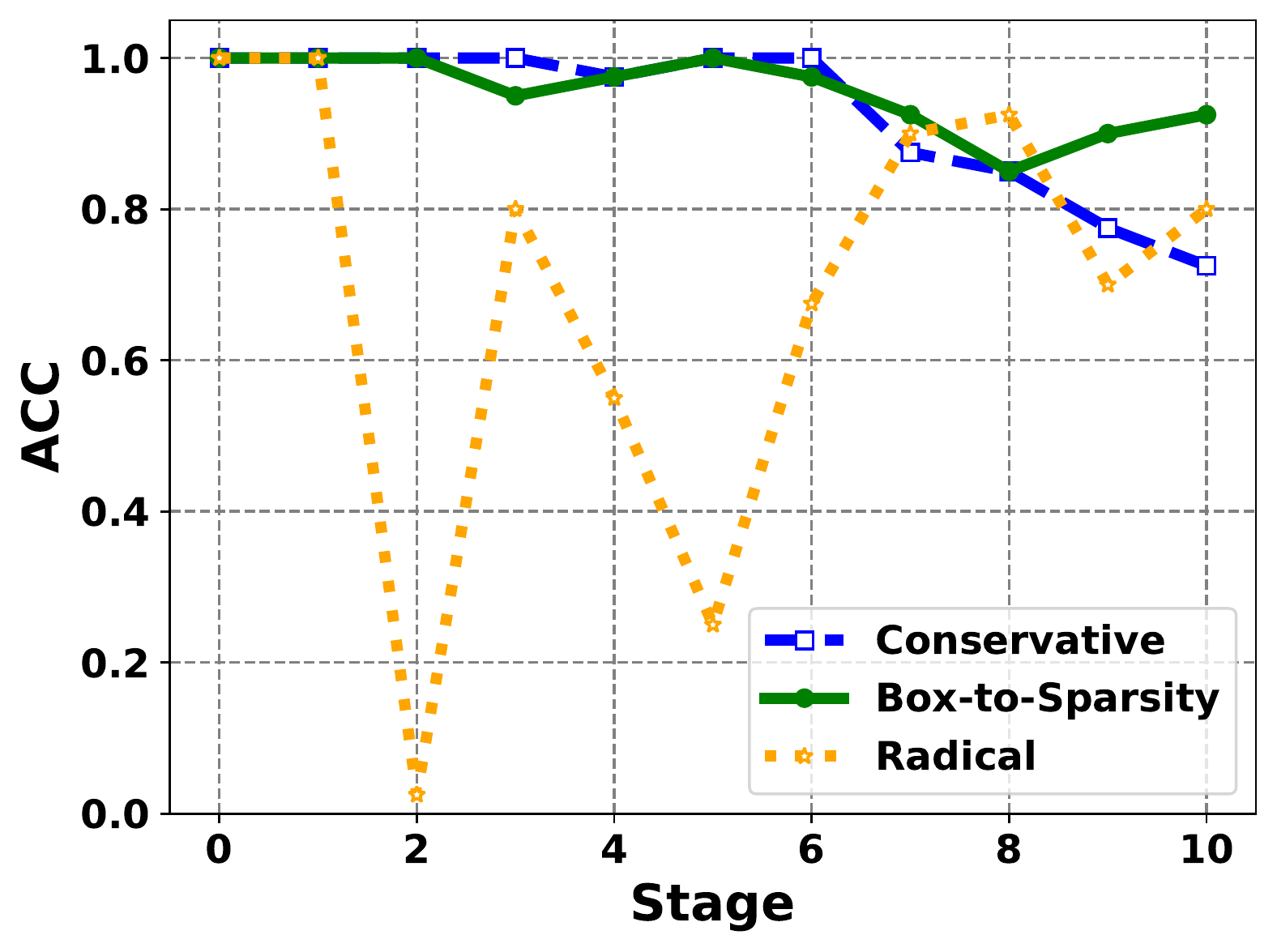}
        \caption{}
        \label{fig:schedule_comparison_clean}
    \end{subfigure}
    \caption{Comparison among different budget scheduling strategies. (a): infected models by BadNets, and (b): clean models.}
    \vspace{-0.1in}
    \label{fig:budget study}
\end{figure}

%% file: resources/fig_8_relationship-adv-trigger.tex
\begin{figure}[tb]
    \centering
    % \vspace{-0.1in}
    \begin{subfigure}[h]{0.49\linewidth}
        \setlength{\abovecaptionskip}{0.cm}
        \includegraphics[width=\textwidth]{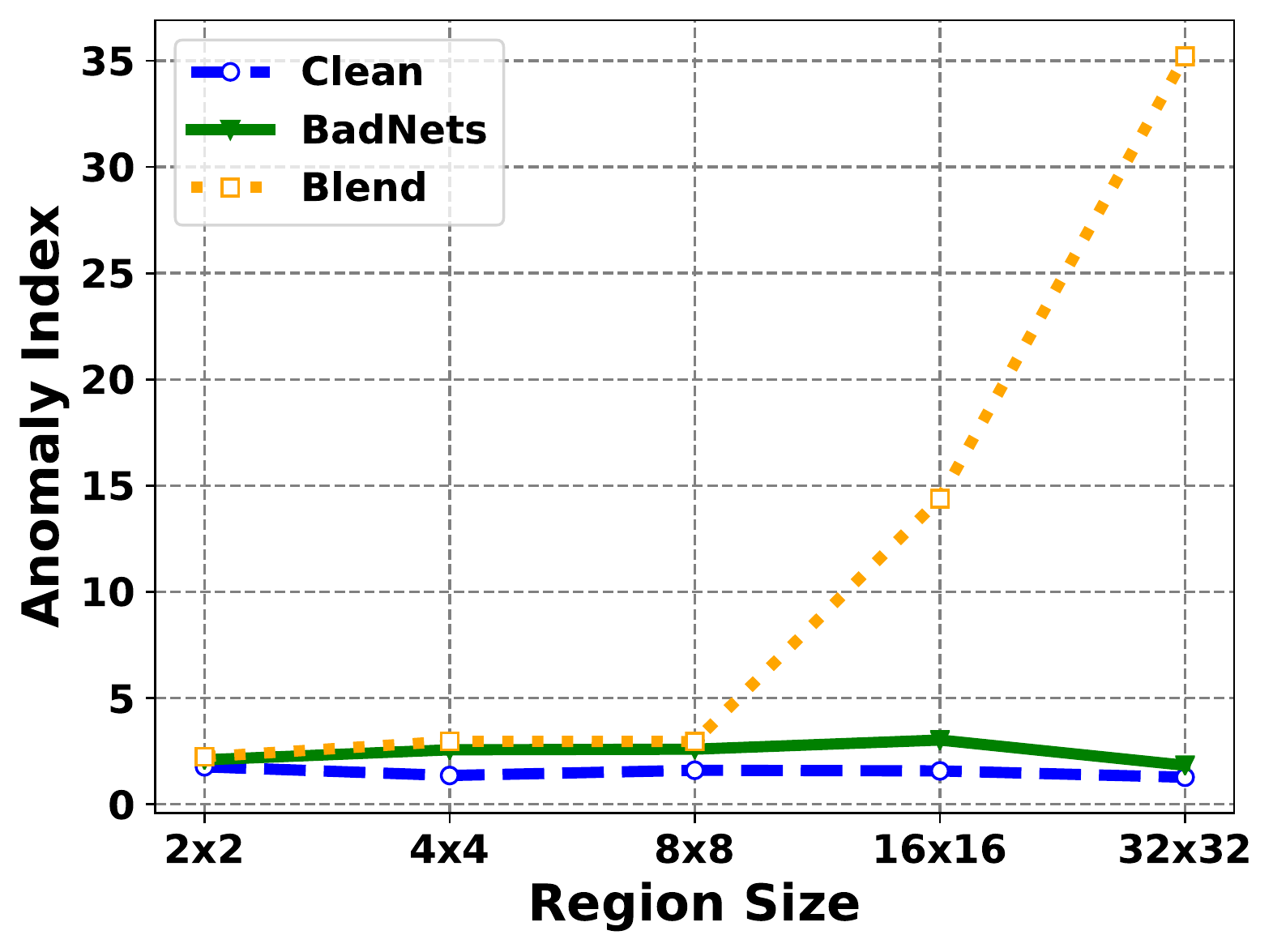}
        \caption{Attack region}
        \label{fig:region}
    \end{subfigure}
    \begin{subfigure}[h]{0.49\linewidth}
        \setlength{\abovecaptionskip}{0.cm}
        \includegraphics[width=\textwidth]{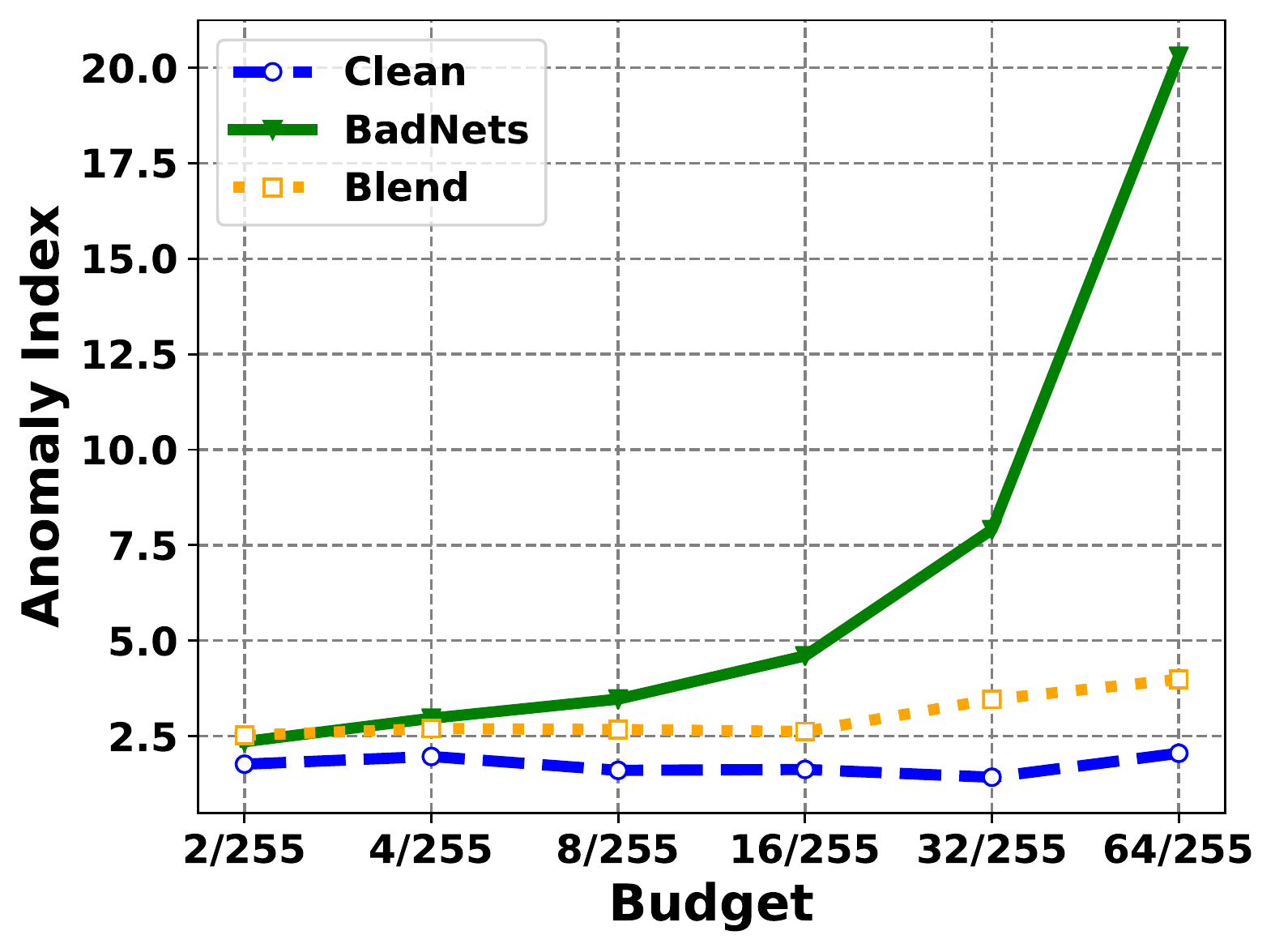}
        \caption{Attack budget}
        \label{fig:budget}
    \end{subfigure}
    \caption{Adversarial perturbations v.s. backdoor triggers studies.}
    \vspace{-0.1in}
    \label{fig:region_budget}
\end{figure}

%% file: resources/tab_2_plugins.tex
\begin{table}[tb]
\vspace{-0.1in}
\tiny
\centering
\caption{Backdoor elimination with A2P. ACC is the accuracy on clean data, while ASR-B represents the attack success rate of backdoor attacks on models. ``Original Trigger'' and ``Random Noise'' indicate fine-tuning with original trigger patterns or noise patterns. ``No Patching'' indicates fine-tuning with clean data.}
\resizebox{0.48\textwidth}{!}{
\begin{tabular}{c|cc|cc} 
\hline
 & \multicolumn{2}{c|}{\textbf{BadNets(\%)}} & \multicolumn{2}{c}{\textbf{Blend(\%)}}  \\ 
\cline{2-5}
  & \textbf{ACC$\uparrow$}   & \textbf{ASR-B$\downarrow$}                & \textbf{ACC$\uparrow$}   & \textbf{ASR-B$\downarrow$ }             \\ 
\hline
\textbf{Before Fine-tune}            & 93.29 & 99.53                & 93.31 & 100                \\
\textbf{No Patching}                 & 92.16 & 99.64                & 93.09 & 100                \\
\textbf{Original Trigger} & 91.26 & 0.16                 & 91.02 & 0.43               \\
\textbf{Random Noise}     & 91.33 & 97.69                & 88.46 & 35.63              \\
\textbf{Reversed Trigger (Ours)} & 91.43 & 1.96                 & 90.53 & 0.7                \\
\hline
\end{tabular}
}
\vspace{-0.1in}
\label{tab:unlearn}
\end{table}

%% file: src/6-conclu.tex
\section{Conclusion}
In this paper, we propose \textit{Adaptive Adversarial Probe} (A2P) framework for backdoor attacks detection in a more complex scenario, where models might be embedded with diverse unforeseen backdoor attacks. Specifically, our A2P adopts a global-to-local probing framework, which adversarially probes images with adaptive regions/budgets using our proposed attention-guided region proposal and box-to-sparsity budget scheduling modules, which could better fit various backdoor triggers of different sizes/transparencies. Extensive experiments demonstrate that our A2P framework outperforms other comparisons by large margins (\textbf{\textit{+12\%}} on \textit{Average Attacks}). In the future, we are interested in proposing an end-to-end learning solution.

%------------------------------------------------------------------------